**Title:**

Data Set Terminology of Deep Learning in Medicine: A Historical Review and Recommendation

**Authors:**

Shannon L. Walston,[1] Hiroshi Seki,[1] Hirotaka Takita,[1] Yasuhito Mitsuyama,[1] Shingo Sato,[2] Akifumi Hagiwara,[3] Rintaro Ito,[4] Shouhei Hanaoka,[5] Yukio Miki,[1] Daiju Ueda[1,6,7]

**Affiliations:**

1) Department of Diagnostic and Interventional Radiology, Graduate School of Medicine, Osaka Metropolitan University, Osaka, Japan
2) Sidney Kimmel Cancer Center at Thomas Jefferson University, Philadelphia, PA, USA
3) Department of Radiology, Juntendo University School of Medicine, Tokyo, Japan
4) Department of Radiology, Nagoya University, Nagoya, Japan
5) Department of Radiology, University of Tokyo Hospital, Tokyo, Japan
6) Department of Artificial Intelligence, Graduate School of Medicine, Osaka Metropolitan University, Osaka, Japan
7) Center for Health Science Innovation, Osaka Metropolitan University, Osaka, Japan

**Abstract:**

Medicine and deep learning-based artificial intelligence (AI) engineering represent two distinct fields each with decades of published history. With such history comes a set of terminology that has a specific way in which it is applied. However, when two distinct fields with overlapping terminology start to collaborate, miscommunication and misunderstandings can occur. This narrative review aims to give historical context for these terms, accentuate the importance of clarity when these terms are used in medical AI contexts, and offer solutions to mitigate misunderstandings by readers from either field. Through an examination of historical documents, including articles, writing guidelines, and textbooks, this review traces the divergent evolution of terms for data sets and their impact. Initially, the discordant interpretations of the word 'validation' in medical and AI contexts are explored. Then the data sets used for AI evaluation are classified, namely random splitting, cross-validation, temporal, geographic, internal, and external sets. The accurate and standardized description of these data sets is crucial for demonstrating the robustness and generalizability of AI applications in medicine. This review clarifies existing literature to provide a comprehensive understanding of these classifications and their implications in AI evaluation. This review then identifies often misunderstood terms and proposes pragmatic solutions to mitigate terminological confusion. Among these solutions are the use of standardized terminology such as 'training set,' 'validation (or tuning) set,' and 'test set,' and explicit definition of data set splitting terminologies in each medical AI research publication. This review aspires to enhance the precision of communication in medical AI, thereby fostering more effective and transparent research methodologies in this interdisciplinary field.

**Introduction**

The integration of deep learning, a subset of artificial intelligence (AI), into medical research has revolutionized the landscape of medical technology, with applications spanning from medical imaging to predictive modeling and clinical decision support.[1–16] However, this integration has not been without its challenges, particularly in the realm of data set terminology standardization. Inconsistencies in terminology across medical and deep learning research have the potential to hinder effective communication, reproducibility, and generalizability of research findings. This review article aims to provide a comprehensive overview of the historical development of data set terminology in both fields, highlight the challenges arising from terminological discrepancies, and propose recommendations for standardization to facilitate more effective collaboration and advancement in medical deep learning research.

The motivation for this review stems from the recognition of the complex interplay of methodologies and terminologies in the evolution of medical deep learning. As the initially distinct vocabularies and conceptual frameworks of medical and AI research have increasingly overlapped, terminological ambiguities have emerged, particularly in the context of data set descriptions. This is notably evident in the use of the term 'validation' and in the approaches to data set categorization.[17] These ambiguities have practical implications for the design, interpretation, and communication of deep learning research in medicine. Our review contributes to the field by providing a comprehensive analysis of the historical development of data set terminology in both medical and AI research, identifying key sources of confusion, and proposing recommendations for standardization. By addressing these terminological challenges, we aim to facilitate more effective interdisciplinary collaboration and promote the development of robust, generalizable, and reproducible deep learning applications in medicine.

In this narrative review we navigate the intricate landscape of terminological discrepancies between the fields of medicine and deep learning, dissecting and elucidating these terminologies. We offer a clear perspective on their historical development, current usage, and the implications of conflicting usages within the realm of deep learning in medicine. This discussion is pivotal in highlighting the potential for confusion and miscommunication that arises when these two disciplines intersect, especially noteworthy in the context of high-stakes deep learning applications in healthcare. Moreover, we seek to clarify the various terms to engender robustness and generalizability of future deep learning applications in medicine. This review, therefore, is not just an exercise in linguistic clarity but a necessary step towards methodological precision and more transparent, reproducible research practices in a rapidly evolving field.

**Historical overview of data set terminology in AI research**

Artificial intelligence, machine learning, and deep learning

The journey of AI research begins with its foundational concepts: AI itself, followed by machine learning, and then deep learning (Figure 1). AI, an idea that took shape in the mid-20th century,[18,19] refers to the broad science of mimicking human abilities. Machine learning,[20] which emerged in the 1980s, is a subset of AI that involves the development of algorithms that can learn from data and make predictions or decisions. Deep learning,[21,22] a 21st century development, is a subset of machine learning that uses layered artificial neural networks, inspired by the structure and function of the human brain, to analyze various factors in large amounts of data. These networks have been central to many breakthroughs in AI, adapting and evolving over time to handle complex and voluminous data sets.

Overview of the terminology history

Among the myriad of advancements in deep learning, there is an equally vast array of terms authors used to describe their work. The systematization of this knowledge by Ripley in 1996 stands out,[23] especially concerning data set terminology (Table 1 and Figure 2). Ripley's work lays down a structured framework for understanding and utilizing data sets in neural network models. This framework is fundamentally categorized into training sets, validation sets, and test sets. Ripley's definitions are as follows:

*Training set: A set of examples used for learning, that is, to fit the parameters of the classifier.*

*Validation set: A set of examples used to tune the parameters of a classifier, for example, to choose the number of hidden units in a neural network.*

*Test set: A set of examples used only to assess the performance of a fully-specified classifier.*

The training set is the first data the model is exposed to and determines the model parameter settings. This set is critical for developing accurate, fair, and generalizable models. The validation set, on the other hand, plays a vital role in tuning those parameters (hyperparameter tuning), preventing overfitting, and guiding the selection of the model's complexity, such as the number of hidden units in a neural network. The ground truth, or expected answer, is input for both of these data sets when developing supervised models. In contrast, the test set hides the ground truth from the model. The test set is used as a final evaluation tool to assess the performance of the model after it has been trained and validated, ensuring that model predictions are as accurate and generalizable as possible.

In the broader realm of deep learning, Ripley's categorization of data sets into training, validation, and test sets is not a novel framework but rather a reflection of practices that have become standard in the field.

These definitions and their corresponding roles have been universally recognized and employed in various deep learning applications. The training set's function as the cornerstone for model learning, the validation set's crucial role in parameter tuning and model refinement, and the test set's importance as a benchmark for performance evaluation are principles deeply ingrained in deep learning methodology. This widespread adoption underlines the effectiveness and essential nature of these classifications in developing robust, efficient, and reliable deep learning systems. Ripley's articulation of these concepts, therefore, echoes a widely accepted and practiced approach within the original domain of deep learning, serving as a critical link in translating these established methodologies to the evolving landscape of medical deep learning.

**Data set terminology in medicine**

Overview of the terminology history

Traditional modeling research in medicine primarily focused on the reliability and validity of the data used to develop a model. In contrast to its use in deep learning research, the term 'validation' referred to testing the developed model without altering any internal parameters. After all, traditional modeling had no internal parameters to adjust; rather they usually used hand-crafted features. As we delve into the history of data set descriptions in medical modeling research guidelines, we observe a gradual, though inconsistent, shift in terminology towards that established in AI research (Table 1 and Figure 2).

Medical guidelines reflect the terminology common at the time of their publication, as well as defining any new terms that researchers may not be familiar with. The original 2003 STARD (Standards for Reporting Diagnostic accuracy studies) guideline,[24] a cornerstone in diagnostic study reporting, reflects the common focus on the end-point 'validation' of diagnostic tools rather than development. The initial QUADAS (Quality Assessment of Diagnostic Accuracy Studies) framework,[25] used similar terminology to STARD as it was developed as a tool for reviewers to check that authors had followed the STARD checklist requirements. Its successor, QUADAS-2,[26] retained the terminology of the original QUADAS, continuing to refer to the 'test' being studied or validated without a term for the different subsets of data that might be used to validate it. The STARD guideline[27] was updated and expanded in 2015, and included an acknowledgement that other guidelines may be more suitable for "more specific forms of [index] tests… such as… multivariable prediction models" than STARD; an indication that model development techniques, and thus the terminology in the field, were beginning to change.

The guideline referred to in the STARD update was the TRIPOD (Transparent Reporting of a multivariable prediction model for Individual Prognosis Or Diagnosis) guideline.[28] This guideline marked a significant shift by adopting terms such as 'model' for the method being studied, 'development' for designing the model, and 'validation' for evaluating the model, with a distinction between 'internal validation' and 'external validation.'[29] This new lexicon acknowledged and standardized the growing practice of developing more complex models based on multiple predictors and evaluating them using several groups of participants in a single study. The emphasis on 'internal validation' and 'external validation' reflects a growing awareness of the need to rigorously evaluate predictive models both within the study context and in external settings. Similar to TRIPOD, the 2019 tool PROBAST (Prediction model Risk Of Bias ASsessment Tool) further underscored the importance of 'external validation'.[30] The successor to TRIPOD, the TRIPOD+AI guideline, opted to use 'evaluation' instead of using the more confusing 'validation'.[31]

Finally, building on the work by Erickson, et al.,[32] Park, et al.,[33] and Bluemke, et al.,[34] the 2020 CLAIM (Checklist for Artificial Intelligence in Medical Imaging) guidelines were the first to explicitly acknowledge the deep learning-centric terminology of model "training, validation ("tuning"), and testing data partitions."[35] The final checklist text uses 'validation or testing' together to bridge the gap in understanding between medical and AI research fields. This inclusion is a clear acknowledgement of the need to harmonize data set terminologies to ensure clarity and consistency across these converging domains. Future guidelines designed specifically for medical AI development[36,37] will likely follow CLAIM by using the deep learning-centric terminology of 'training,' 'validation,' and 'test.'

The progression in data set terminology within medicine, from the traditional testing-centric language to incorporating model performance validation prior to testing, illustrates an ongoing adaptation to the changing landscape of medical research. This recognition of the need for clear, standardized terminologies that can bridge the gap between traditional medical research methodologies and the emerging practices in medical deep learning ensures the continued integrity and efficacy of research in this interdisciplinary field.

**Toward standardization for data set terminology of deep learning in medicine**

Terminology conflict between medicine and deep learning

The term 'validation' epitomizes the terminological conflict between traditional medicine and deep learning research, as described above. In medical research, 'validation' often refers to the process of confirming the applicability and accuracy of a diagnostic tool in a clinical setting, which is termed 'testing' in deep learning research. While in deep learning research, 'validation' denotes a phase in the model development process, sandwiched between training and testing. This divergence in meaning can lead to significant misunderstandings, particularly in interdisciplinary collaborations where each field brings its own language and interpretive frameworks.

Examples of inconsistency in famous deep learning studies in medicine

Even in the first articles on deep learning published in prestigious medical journals, confusion regarding data set terminology can be observed (Table 1 and Figure 2). The 2016 paper by Gulshan, et al. in the Journal of the American Medical Association (JAMA),[38] titled "Development and Validation of a Deep Learning Algorithm for Detection of Diabetic Retinopathy in Retinal Fundus Photographs," is one example. The terms 'tuning set' and 'validation set' are used for what are typically known in deep learning research as the 'validation set' and 'test set,' respectively. Although most of the authors are AI specialists who use the deep learning-centric data set terms when writing for other journals,[39] they chose to use medical research-centric data set terms when publishing in JAMA. Researchers of deep learning in medicine may be confused to find that a study reproducing that 2016 Gulshan, et al. work[40] used the terms 'validation set' and 'test set,' specifically noting the terminology change from the study they were reproducing. Similarly, the 2018 study by Chilamkurthy, et al. in The Lancet,[41] "Deep learning algorithms for detection of critical findings in head CT scans: a retrospective study," utilizes the terms 'development' and 'validation' rather than 'training/validation' and 'test,' like the same authors use in non-medical journals.[42] Authors of review articles face this inconsistency head-on as they must interpret the various terms used in each study, and then choose which terminology to use in their work.[17] This inconsistent terminology between journals could easily confuse researchers.

We see a shift towards deep learning research terminology after these works, exemplified by the 2020 study by Milea, et al. in the New England Journal of Medicine,[43] entitled "Artificial Intelligence to Detect Papilledema from Ocular Fundus Photographs." This work utilizes the traditional deep learning nomenclature, employing the terms 'training set,' 'validation set,' and 'testing set.' Notably, they clearly defined each data set term in the appendix. In the years since, no set terminological standard has been adopted, though some journals endorse one or more of the guidelines.

Standardization for data set terminology

Some researchers began advocating for standardization to encourage clarity. The 2019 study by Xiaoxuan Liu, et al. in The Lancet Digital Health,[17] "A comparison of deep learning performance against health-care professionals in detecting diseases from medical imaging: a systematic review and meta-analysis," advocated for the medical research-centric terminology where 'validation test set' is used for the final testing, and developmental sets are termed 'training' or 'tuning' sets. By 2020, several additional works discussing the inconsistencies between deep learning research and medical research terminology had been widely disseminated. These works, including the 2019 studies by Ting, et al.[44] and Yun Liu, et al.,[45] encouraged the use of terminology common in deep learning research.

We support two changes to ensure the seamless flow of ideas between both fields (Table 2). First and foremost, the use of clear, standardized terms like 'training set,' 'validation set,' and 'test set' should be encouraged.[23,46–50] These terms have already gained wide acceptance in the deep learning field and can provide a common ground for communication. An alternative could be the adoption of 'tuning set' in place of 'validation set,' which might offer a more intuitive understanding for medical professionals regarding the role of this data set in optimizing the deep learning model.[35,51] Second is the explicit definition of these terms in each study. By clearly articulating what each term means in the context of a specific research project, researchers can significantly reduce the risk of misinterpretation and confusion.[43,52] Future collaborative research demands a heightened focus on clarity and transparency in research methodologies to ensure that all collaborators from all disciplinary backgrounds have a mutual understanding of the terms and processes involved.

The harmonization of data set terminology between medical and deep learning research is critical for advancing the field. By adopting standardized terminologies and ensuring their clear definition in research publications, we can foster more effective communication and collaboration across disciplines. This approach will not only facilitate a better understanding of deep learning applications in medicine but also contribute to their more ethical and effective deployment in clinical settings.

**Categories of test sets**

Overview of test set categories

In the field of AI, especially within the context of medical deep learning, the nature of the data set used to test the model can significantly influence the performance, applicability, and generalizability of deep learning tools.[53] Similarly to 'validation,' the usage of terminology for the testing set is not standardized and often unclear from the manuscript. This section describes the various categories of test sets commonly employed in deep learning evaluation, decoding their unique characteristics and implications for deep learning research (Table 3).

The classification of data sets in deep learning evaluation is typically based on their source, composition, and intended use in the development lifecycle. The data source may be internal or external, relative to the training data. Internal data sets are derived from the same pool as the training data, while external data sets are completely unrelated to the training data. The primary composition categories for internal data sets include random splitting, cross-validation, and leave-one-out methods. For external data sets, the main categories are temporal and geographic (both national and international) sets. Random splitting involves dividing a data set into separate sets randomly, while cross-validation partitions the data set into several subsets for training and testing. The leave-one-out method is a specific form of cross-validation where each data point is used once as the test set. Temporal sets are collected at a different time from the training data, while geographic sets are split based on where the data was collected. This classification is important for understanding how the test sets are structured and how they might impact the outcomes of deep learning models. Each type of test set has unique characteristics that influence the model's performance, generalizability, and potential real-world applicability.

Internal testing: random splitting, cross-validation, and leave-one-out test sets

Internal testing refers to testing using a test set that was split from the same pool as that used for training data. Thus, internal test sets have demographics very similar to the training data, but the model has not seen that specific data. When using internal data, the way the data is split affects the time and computational power required to train the model, as well as the statistical power of the evaluation results. Common internal data splitting methods include random splitting, cross-validation, and leave-one-out methods.

Random splitting involves dividing a data set into separate sets (usually 'training,' 'validation,' and 'testing') randomly. This method is the simplest to ensure that models are not biased towards specific subsets of data. Cross-validation involves partitioning the data set into several subsets, often five or more, then training the model on some of the subsets while keeping the others separate for testing. [54–57] Confusingly, 'validation' is used here, but in many cases, the sets divided by this cross-validation fall within the scope of internal testing. The subsets are often divided using random splitting, though they may be curated to ensure equal distribution of certain characteristics if necessary. This approach is more robust than a single random split and particularly useful for assessing model performance across different data samples because the model is tested on each splitting combination and the results are merged into one final performance score. The leave-one-out method is a specific form of cross-validation where each data point is used once as the test set while the rest form the training set, offering a thorough evaluation of the model at the expense of computational intensity.

External testing: temporal and geographic test sets

External testing refers to testing using data completely unrelated to that used for training and internal testing. External testing on various potential sources of data is necessary in order to determine if the deep learning model erroneously learned some shortcut exclusive to the internal set rather than the intended target features. External test sets may differ from training data temporally or geographically.

Temporal and geographic test sets are distinct from the training set based on when or where they were collected. Temporal sets are those collected at a different time, typically with training on historical data and testing on recent or future data.[58–61] While there is some debate on whether temporal test sets should be considered truly external, guidelines such as PROBAST, TRIPOD+AI, and work by Park et al. define them as external test sets.[30,31,33] This is because the temporal test data was not used to train or validate the model and is temporally distinct from the training data. Temporal test sets are crucial in applications where temporal dynamics play a significant role, such as in predictive medical models.

Geographic sets are split based on where the data was collected. The use of national geographic sets involves data sourced from specific regions or facilities within a country. These data sets are instrumental in assessing the deep learning model's performance within a national population, highlighting issues such as regional biases or applicability.[62] International geographic sets expand this concept, encompassing data from multiple countries. External sets are vital for evaluating the generalizability and scalability of deep learning models across diverse settings.

**Conclusion**

The continued evolution of medical research and AI terminology into medical AI terminology underscores the dynamic nature of this domain. We support the use of terminology that reduces the risk of miscommunication of ideas and encourage the definition of often miscommunicated terms within each text to improve understanding between both fields.

The disparities in data set terminology between traditional medical and deep learning research, specifically illustrated in the varied interpretations of the term 'validation,' demonstrate the critical need for a unified language. This terminological alignment is not merely a matter of semantics but a foundational requirement for effective communication and collaboration across disciplines. Our review of prominent deep learning studies in medicine reveals the tangible impact of these discrepancies, where misunderstandings can lead to misinterpretation of research findings and methodologies. Such scenarios emphasize the importance of developing a standardized lexicon that resonates with both medical practitioners and deep learning researchers.

Our recommendations for the harmonization of data set terminology in medical deep learning research are anchored in the pursuit of clarity and precision. The adoption of standardized terms such as 'training set,' 'validation (or tuning) set,' and 'test set,' alongside explicit definitions in research publications, is paramount. Additionally, test set terminology should also be used in an appropriate manner to show the model performance correctly. This approach ensures that the methodologies and results of deep learning applications in healthcare are accurately understood and effectively utilized.

Looking forward, the need for continuous dialogue and collaboration between the fields of medicine and deep learning is evident. As deep learning continues to evolve and permeate various aspects of medicine, the integration of a shared language and understanding of data set terminology will be instrumental in advancing the field of medical deep learning. This harmonization is not just beneficial for researchers and practitioners but is also crucial in ensuring that deep learning applications in medicine are ethical, effective, and reliable.

Data availability statement

The authors confirm that the data supporting the findings of this study are available within the article.

**Tables**
**Table 1. Terminology usage history**

|  | Year | Type of publication | Terminology | | |
|---|---|---|---|---|---|
|  |  |  | Model training | Model parameter tuning | Model evaluation |
| Medical field |  |  |  |  |  |
| STARD[24,27] | 2003/2015 | Guideline | - | - | test |
| TRIPOD[28] | 2015 | Guideline | develop | develop | validation |
| The first original DL article in JAMA[38] | 2016 | Original article | develop | develop | validation |
| The first original DL article in The Lancet[41] | 2018 | Original article | develop | develop | validation |
| PROBAST[30] | 2019 | Guideline | develop | develop | validation |
| CLAIM[35] | 2020 | Guideline | train | validation (tuning) | test |
| The first original DL article in The NEJM[43] | 2020 | Original article | train | validation (five-fold) | test |
| TRIPOD+AI[31] | 2024 | Guideline | develop | develop | evaluation |
| Engineering field |  |  |  |  |  |
| Pattern Recognition and Neural Networks[23] | 1996 | Book | train | validation | test |

DL; deep learning.

**Table 2. Recommendations for data set terms**

| Recommendation | Description |
|---|---|
| Use of standardized data set terms | Encourage the adoption of terms such as 'training set,' 'validation set,' and 'test set.' These are widely accepted in the deep learning field and facilitate clear communication. An alternative term, 'tuning set,' could be used for 'validation set' to provide an intuitive understanding for medical professionals. |
| Explicit definition of data set terms | Each study should clearly define the terms 'training set,' 'validation set' (or 'tuning set'), and 'test set' in its context. This reduces the risk of misinterpretation and confusion by explicitly stating what each term means within the specific research project. |

**Table 3. Test set variations**

| Type of Testing | Description | Key Characteristics |
| --- | --- | --- |
| Internal Testing | Testing using a test set split from the same pool as training data. | ● Similar demographics to training data. |
| Random Splitting | Divides the data set randomly. | ● Simplest method.<br>● Prevents model bias towards specific data subsets. |
| Cross-validation | Partitioning the data set into several subsets, training on some while testing on others. | ● More robust than random splitting.<br>● Assesses performance across different samples.<br>● Computationally intensive.<br>● Results from each subset are combined. |
| Leave-One-Out | Each data point is used once as the test set, the rest as the training set. | ● Thorough evaluation.<br>● Computationally intensive.<br>● Results from each subset are combined. |
| External Testing | Testing using data unrelated to that used for training and internal testing. | ● Different demographics from training data.<br>● Determines if the model learned features exclusive to the internal set. |
| Temporal Set | Collected at different times than the training data. | ● Assesses how well the model predicts using recent/future data.<br>● Important where time dynamics are significant. |
| Geographic Set | Based on where the data was collected, either nationally or internationally. | ● Evaluates performance in a geographically distinct population.<br>● Addresses regional biases.<br>● Tests model generalizability and scalability. |

**Figure legends**
**Figure 1. History of artificial intelligence, machine learning, and deep learning**

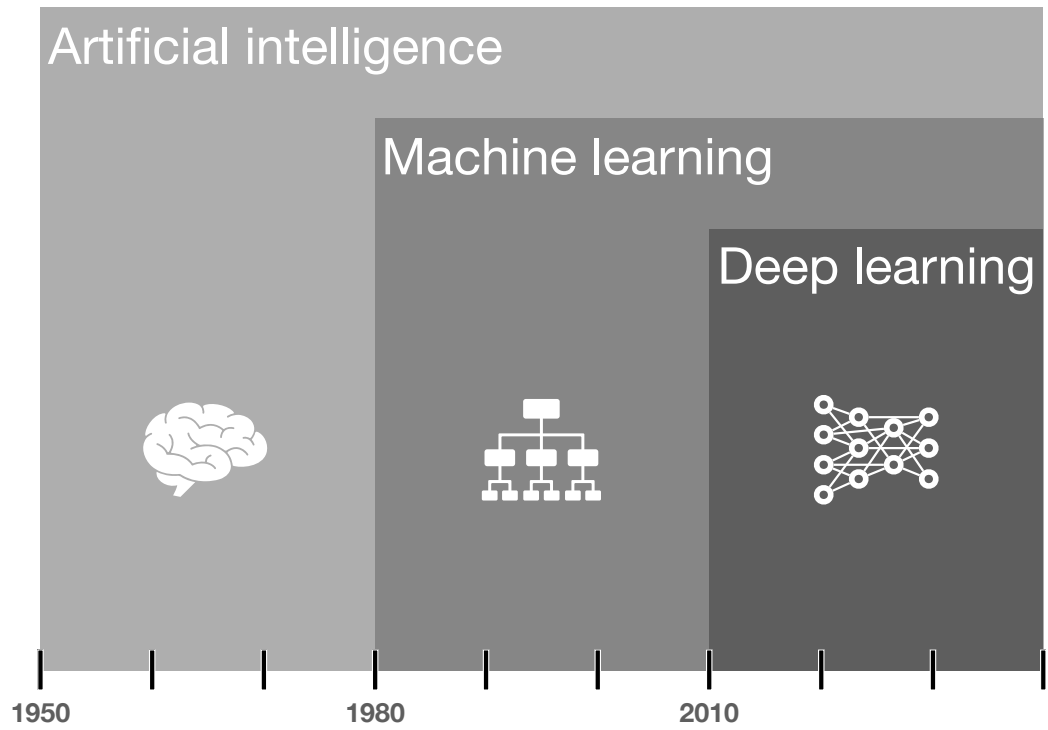

**Figure 2. Timeline of terminological guidance in the medical and engineering fields**

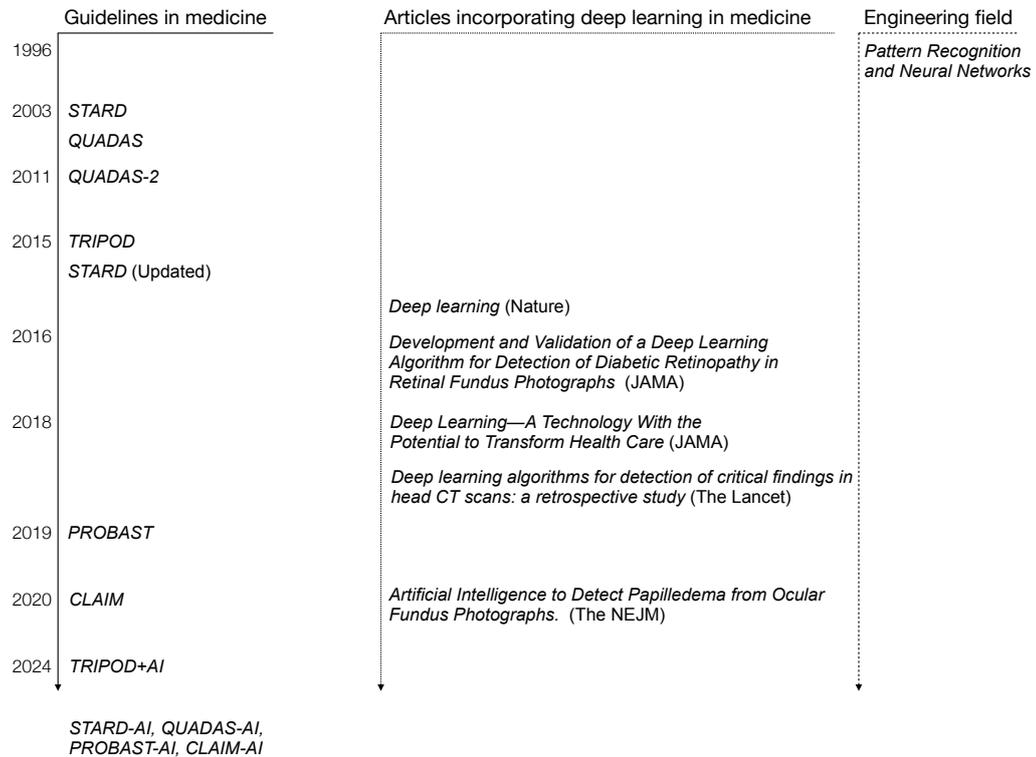

This figure presents a timeline from 1996 to 2020 highlighting the development of guidelines in medicine and the corresponding incorporation of AI in medical research, as well as its connection to the engineering field of pattern recognition and neural networks. The left column lists the introduction of various medical modeling guidelines, such as STARD,[24,27] QUADAS,[25,26] TRIPOD,[28] PROBAST,[30] CLAIM,[35], and TRIPOD+AI [31] including their updates. The middle column shows significant articles from prominent medical journals like Nature,[21] JAMA,[38,22], The Lancet,[41] and The NEJM[43] that have incorporated deep learning. These articles mark milestones in the application of deep learning in medicine. The right column categorizes these advancements under the broader engineering field, emphasizing the interdisciplinary nature of modern medical research.

**Figure 3. Data partitioning schemes**

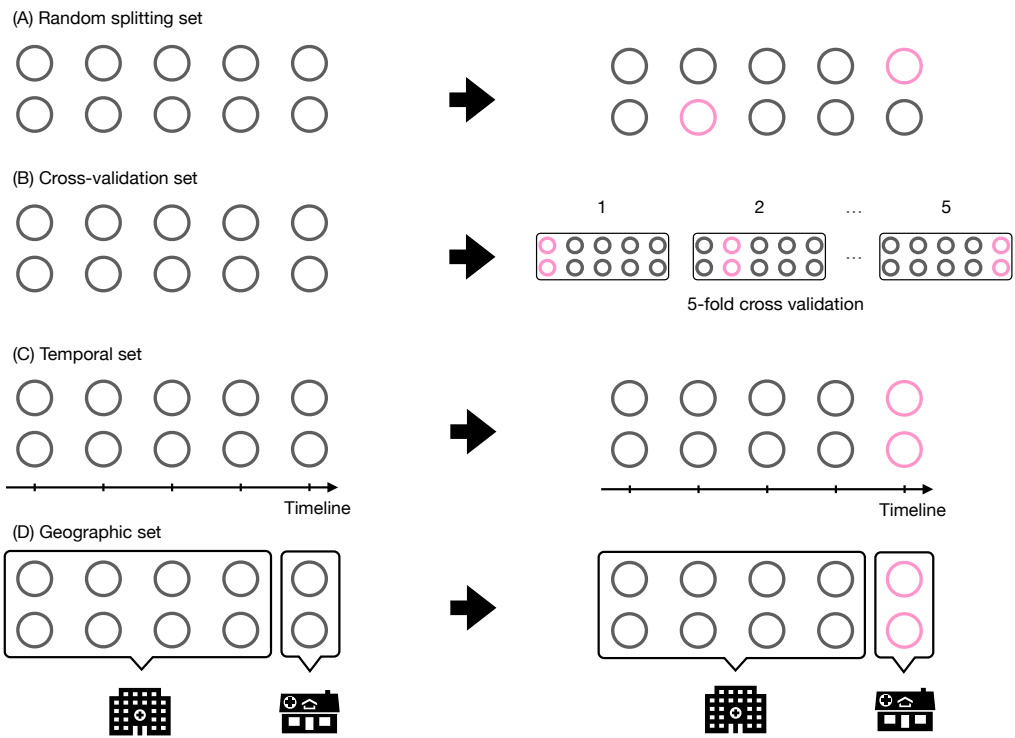

(A) Random Splitting Set
In this method, the data set is randomly divided, with the pink circles representing the portion of data allocated as the test set. The test set is randomly sampled from the entire data set.

(B) Cross-Validation Set
This illustrates a 5-fold cross-validation technique. The data set is split into five equal parts, with each part taking turns being the test set (pink). This process is repeated five times and the results are averaged, with each part serving as the test set once.

(C) Temporal Set
The data set is divided according to time, with the test set (pink) consisting of the most recent data. It is not necessary to use the most recent data for the test set, but if the goal is to create a model that predicts recent data, it is most practical to use the most recent data for the test set. This is suitable for time-series data where the goal is to predict future events based on past data.

(D) Geographic Set
The data set is divided based on geographic location. The test set (pink) comprises data from one of several locations, indicated by map pin symbols. This approach is often used when model performance needs to be evaluated on data from distinct geographic regions.